\documentclass[11pt]{article}
\usepackage{hyphenat}

\usepackage[preprint]{acl}  

\usepackage{times}
\usepackage{latexsym}

\usepackage[T1]{fontenc}
\usepackage[utf8]{inputenc}

\usepackage{microtype}

\usepackage{inconsolata}
\usepackage{microtype}

\usepackage{graphicx}
\usepackage{tcolorbox}
\usepackage{amsmath}
\usepackage{amssymb}
\usepackage{graphicx}
\usepackage{booktabs}
\usepackage{algorithm}
\usepackage{algpseudocode}
\usepackage{hyperref}
\usepackage{times}
\usepackage{tabularx}
\usepackage{natbib}
\usepackage{hyperref}
\usepackage[bottom]{footmisc}

\newtheorem{theorem}{Theorem}
\newtheorem{definition}{Definition}

\newenvironment{proof}{\paragraph{Proof:}}{\hfill$\square$}

\title{\textbf{Debate, Deliberate, Decide (D3): A Cost-Aware Adversarial Framework for Reliable and Interpretable LLM Evaluation}}

\author{
  Abir Harrasse$^{1}$, \quad Chaithanya Bandi$^{1,2}$, \quad Hari Bandi$^{3}$ \\
  $^{1}$Martian \quad $^{2}$NUS \quad $^{3}$MIT\\
  \texttt{\{abir.harrasse@emines.um6p.ma, bizchaba@nus.edu.sg\}}
}

\begin{document}

\maketitle
\begin{abstract}
The evaluation of Large Language Models (LLMs) remains challenging due to inconsistency, bias, and the absence of
transparent decision criteria in automated judging. We present \textbf{Debate, Deliberate, Decide (D3)}, a cost-aware,
adversarial multi-agent framework that orchestrates structured debate among role-specialized agents (advocates, a judge,
and an optional jury) to produce reliable and interpretable evaluations. D3 instantiates two complementary protocols:
(1) \emph{Multi-Advocate One-Round Evaluation (MORE)}, which elicits $k$ parallel defenses per answer to amplify
signal via diverse advocacy, and (2) \emph{Single-Advocate Multi-Round Evaluation (SAMRE)} with \emph{budgeted stopping},
which iteratively refines arguments under an explicit token budget and convergence checks.

We develop a probabilistic model of score gaps that (i) characterizes reliability and convergence under iterative debate
and (ii) explains the separation gains from parallel advocacy. Under mild assumptions, the posterior distribution of the
round-$r$ gap concentrates around the true difference and the probability of mis-ranking vanishes; moreover, aggregating
across $k$ advocates provably increases expected score separation. We complement theory with a rigorous experimental
suite across \textsc{MT-Bench} \cite{zheng2023judging}, \textsc{AlignBench} \cite{liu2024alignbench}, and \textsc{AUTO-J} \cite{li2023generativejudgeevaluatingalignment}, showing state-of-the-art agreement with human
judgments (accuracy and Cohen's $\kappa$), reduced positional and verbosity biases via anonymization and role
diversification, and a favorable cost--accuracy frontier enabled by budgeted stopping. Ablations and qualitative analyses
isolate the contributions of debate, aggregation, and anonymity.

Together, these results establish D3 \footnote{Code Available at: \url{https://github.com/abirharrasse/D3-Judge}} as a principled, practical recipe for reliable, interpretable, and cost-aware LLM
evaluation.
\end{abstract}

\begin{figure*}[h]
    \centering
    \includegraphics[width=\linewidth]{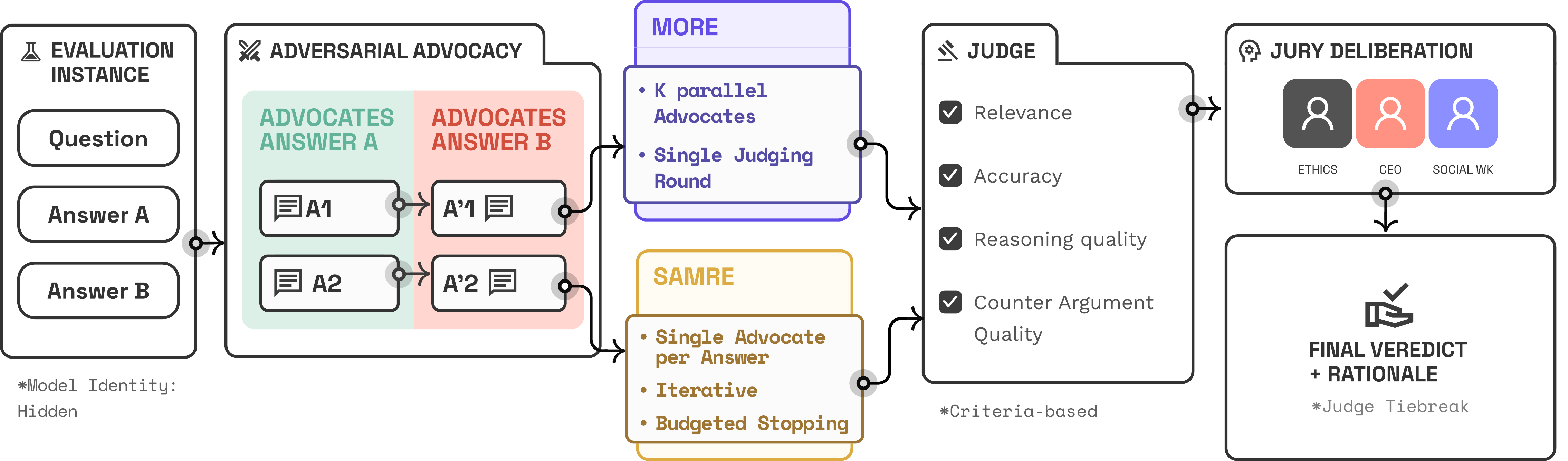}
    \caption{The D3 system routes pairwise evaluations through structured adversarial debate. Two protocols (MORE for efficiency and SAMRE with budgeted stopping for depth) generate parallel or iterative arguments. A Judge provides criteria-based scoring to guide refinement. A diverse jury panel independently evaluates the anonymized debate transcript and renders a final verdict, with ties broken by the Judge's cumulative score.}
    \label{fig:main_figure}
\end{figure*}

\section{Introduction}
The rapid proliferation of Large Language Models (LLMs) \cite{brown2020language} has created significant challenges in evaluating their increasingly complex capabilities, particularly in open-ended generation tasks \cite{celikyilmaz2020evaluation}. Traditional automated metrics such as BLEU \cite{papineni2002bleu} and ROUGE \cite{lin2004rouge} fail to capture semantic coherence, factual accuracy, or alignment with human values \cite{callison2006bleu, mathur2020tangled}. Consequently, human evaluation remains the gold standard \cite{howcroft2020confusion}, but its high cost, slow turnaround, and inherent subjectivity make it impractical for iterative development cycles \cite{liang2022helm}. This has motivated the ``LLM-as-a-Judge'' paradigm \cite{zheng2023judging, kim2023prometheus}, where powerful LLMs evaluate other models' outputs, showing promising alignment with human preferences. Such approaches are critical for training helpful assistants via reinforcement learning from human feedback \cite{bai2022training, christiano2017deep, ziegler2020finetuning} and building aligned language assistants \cite{askell2021general}.
However, single LLM judges are susceptible to positional bias, verbosity bias, and self-enhancement bias \cite{wang2023fair, mehrabi2021survey}. Multi-agent approaches like ChatEval \cite{chan2023chateval} and PRD \cite{li2023prd} mitigate these through diverse personas and peer discussion, yet critical gaps remain: insufficient empirical rigor across diverse benchmarks, lack of dedicated bias auditing methodologies, and cost-agnostic designs despite computational expense being a fundamental adoption barrier.
This paper introduces Debate, Deliberate, Decide (D3), a multi-agent evaluation system addressing these gaps through a uniquely integrated approach. Our contributions are:
\begin{enumerate}
\item \textbf{Courtroom-inspired architecture} with explicit Advocate/Judge/Juror role specialization and structured adversarial debate protocols.
\item \textbf{Dual cost-aware protocols}: MORE for parallelized efficiency and SAMRE with budgeted stopping for iterative depth, offering the first explicit cost-accuracy trade-off menu for practitioners.
\item \textbf{Theoretical grounding} via probabilistic convergence models and formal separation proofs that justify design choices and stopping criteria.
\item \textbf{Systematic bias auditing methodology} that quantifies and demonstrates robustness to positional and self-enhancement biases through controlled experiments.
\item \textbf{Rigorous multi-benchmark validation} across MT-Bench, AlignBench, and AUTO-J against strong baselines (ChatEval, PRD, PandaLM).
\end{enumerate}
By combining all five dimensions, D3 provides a scalable, interpretable, and cost-sensitive solution that advances the state of the art in trustworthy LLM evaluation.

\section{The Debate, Deliberate, Decide (D3) Framework}
\subsection{Agent Architecture and Role Specialization}
The D3 framework employs three specialized agent roles, each fulfilled by an LLM guided by specific instructional prompts. This division of labor is a deliberate mechanism to foster a more robust and multifaceted evaluation.

\begin{itemize}
    \item \textbf{Advocates:} These agents are tasked with constructing the most compelling arguments in favor of a specific candidate response. For a given question and two answers, two sets of advocates work independently. Their objective is not to be impartial but to be persuasive, focusing on criteria such as factual accuracy, relevance, depth, and clarity. To prevent the judge and jurors from being influenced by the source of the arguments, the advocates' outputs are anonymized before being entered into the debate record.
    \item \textbf{Judge:} This agent acts as a moderator and facilitator of the debate. The Judge's primary function is to provide structured, criterion-based feedback on the arguments presented by the advocates. It scores each side's defense on a predefined rubric (e.g., Relevance, Accuracy, Reasoning). This scoring serves as a signal for iterative refinement in multi-round debates and as a tie-breaking mechanism in the final decision.
    \item \textbf{Jurors:} The final decision rests with a panel of LLM agents assigned diverse, predefined personas, such as "a retired professor of ethics," "a technology entrepreneur," or "a social worker". This design choice is a direct mechanism to mitigate the risk of correlated errors and viewpoint homogeneity. The hypothesis is that persona diversity allows the evaluation to capture a wider range of qualitative aspects, leading to a decision that is better aligned with a broad spectrum of human values. We validated the robustness of persona selection by testing 50 diverse personas across varying domains; details in Appendix~\ref{subsec:persona_pool}.
\end{itemize}
\subsection{The Adversarial Debate Protocols}
D3 incorporates two distinct protocols to manage the debate, allowing users to select an approach that best fits their needs for speed, cost, and depth of analysis.

\begin{itemize}
    \item \textbf{Multi-Advocate One-Round (MORE):} This protocol is optimized for breadth and efficiency. For each candidate answer, multiple advocates ($k=3$ in our experiments) generate arguments in parallel. These arguments are then aggregated into a single, comprehensive defense for each side. The Judge evaluates these two consolidated defenses in a single round. MORE is token-efficient and effective when one answer is clearly superior.
    \item \textbf{Single-Advocate Multi-Round (SAMRE):} This protocol is designed for depth and iterative refinement. A single advocate for each answer engages in a turn-based debate over multiple rounds. In each round, advocates use the Judge's feedback and their opponent's argument from the previous round to refine their position. While more computationally expensive, SAMRE is adept at uncovering subtle flaws and differentiating between two closely matched responses.
\end{itemize}

To manage the cost of the SAMRE protocol, D3 introduces a \textbf{Budgeted Stopping Rule}. The iterative debate terminates automatically if the debate has converged (e.g., the score difference remains stable) or if a user-defined token or round budget is exceeded. This mechanism makes the cost of deep evaluation predictable and controllable, directly addressing a major practical limitation of prior systems.
\subsection{Deliberation and Aggregation}
The final phase of the D3 process ensures that the verdict is based on a comprehensive review of all evidence generated during the debate.
\begin{enumerate}
    \item \textbf{Transcript Compilation:} Upon conclusion of the debate, a complete, anonymized transcript is compiled, including the original question, candidate answers, all arguments, and all feedback and scores from the Judge.
    \item \textbf{Jury Deliberation:} The full transcript is presented to each member of the Juror panel. Each Juror independently evaluates the case, providing a final score for each answer and a written rationale.
    \item \textbf{Verdict Aggregation:} The final verdict is determined by a majority vote of the jurors. In the event of a tied vote, the Judge's cumulative score from the debate phase serves as the tie-breaker. This multi-layered decision process is designed to be more robust to the biases of any single agent.
\end{enumerate}

\section{Theoretical Framework}

\begin{definition}
\textbf{Gap Distribution and Bayesian Update.}
We model the gap $\delta_r$ at round $r$ as a Beta-distributed random variable. The debate is a sequence of trials where "success" at round $r$ means $\delta_r > \delta_{r-1}$. With prior $\text{Beta}(\alpha_0, \beta_0)$ and $w_r$ cumulative successes up to round $r$, the posterior is:
\[
\delta_r \sim \text{Beta}(\alpha_0 + w_r, \beta_0 + r - w_r)
\]
The expected gap is $\mathbb{E}[\delta_r] = \frac{\alpha_r}{\alpha_r + \beta_r}$ with variance decreasing at rate $O(1/r)$, signifying increasing confidence.
\end{definition}

\begin{theorem}[Probabilistic Convergence]
If the expected gap converges to a true differentiation level $\Delta > 0$, then for any tolerance $\epsilon > 0$:
\[
\lim_{r\to\infty} P(|\delta_r - \Delta| < \epsilon) = 1.
\]
\end{theorem}
\begin{proof}
See Appendix~\ref{app:prob_convergence}.
\end{proof}

\noindent\textbf{Posterior dynamics and concentration.} The round-$r$ gap follows $\delta_r \sim \text{Beta}(\alpha_0 + w_r, \beta_0 + r - w_r)$ with posterior mean \[
\mathbb{E}[\delta_r] = \frac{\alpha_0 + w_r}{\alpha_0 + \beta_0 + r}
\]
and concentration bound \[P(\delta_r \ge 1 - \epsilon) \ge 1 - \frac{4 \cdot \mathrm{Var}(\delta_r)}{\epsilon^2}\].

\begin{theorem}
\textbf{Score-Separation via Parallel Advocacy}
\label{thm:separation}
For $k$ independent defenses $f_{i,j}$ per answer and judge scoring functional $g(\cdot)$:
\[
\begin{split}
\mathbb{E}\!\left[\left|\max_{j} g(f_{1,j}) - \max_{j} g(f_{2,j})\right|\right] \\
> \mathbb{E}\!\left[\left|g(f_1) - g(f_2)\right|\right].
\end{split}
\]
\end{theorem}
\begin{proof}
See Appendix~\ref{app:score_separation}.
\end{proof}

These results formalize how iterative debate concentrates uncertainty while parallel defenses amplify signal.

\subsection{Comparative Analysis of Debate Protocols}
We analyze the theoretical advantages of the multi-advocate (MORE) protocol compared to single-advocate, iterative (SAMRE) approaches. Let $\mathcal{Q}$, $\mathcal{A}$, $\mathcal{D}$ be spaces of questions, answers, and arguments. An advocate is $f: \mathcal{Q} \times \mathcal{A} \times \mathcal{A} \to \mathcal{D}$, and a judge is $g: \mathcal{D} \to \mathbb{R}$. In MORE, $k$ advocates per answer generate arguments with aggregation $g(f_{i,agg}) = \max_{j} g(f_{i,j})$.

\begin{theorem}[Multi-Advocate Superiority]
\label{thm:more}
If superior answer scores stochastically dominate inferior ones, then:
\[
\mathbb{E}[|g(f_{1,agg}) - g(f_{2,agg})|] > \mathbb{E}[|g(f_1) - g(f_2)|].
\]
\end{theorem}
\begin{proof}
See Appendix~\ref{app:score_differentiation}.
\end{proof}

\section{Experimental Design for Rigorous Validation}
\subsection{Benchmarks and Evaluation Tasks}
We evaluate on three benchmarks targeting different LLM capabilities:
\begin{itemize}
\item \textbf{MT-Bench:} 80 multi-turn conversational questions testing general-purpose helpfulness and instruction-following \cite{zheng2023judging}.
\item \textbf{AlignBench:} 683 alignment-focused questions covering helpfulness, harmlessness, and ethical reasoning (professionally translated to English) \cite{liu2024alignbench}.
\item \textbf{AUTO-J:} 58 real-world scenarios with 3,436 pairwise comparisons spanning creative writing, technical explanation, and diverse task domains \cite{li2023generativejudgeevaluatingalignment}.
\end{itemize}

\begin{figure}
    \centering
    \includegraphics[width=\linewidth]{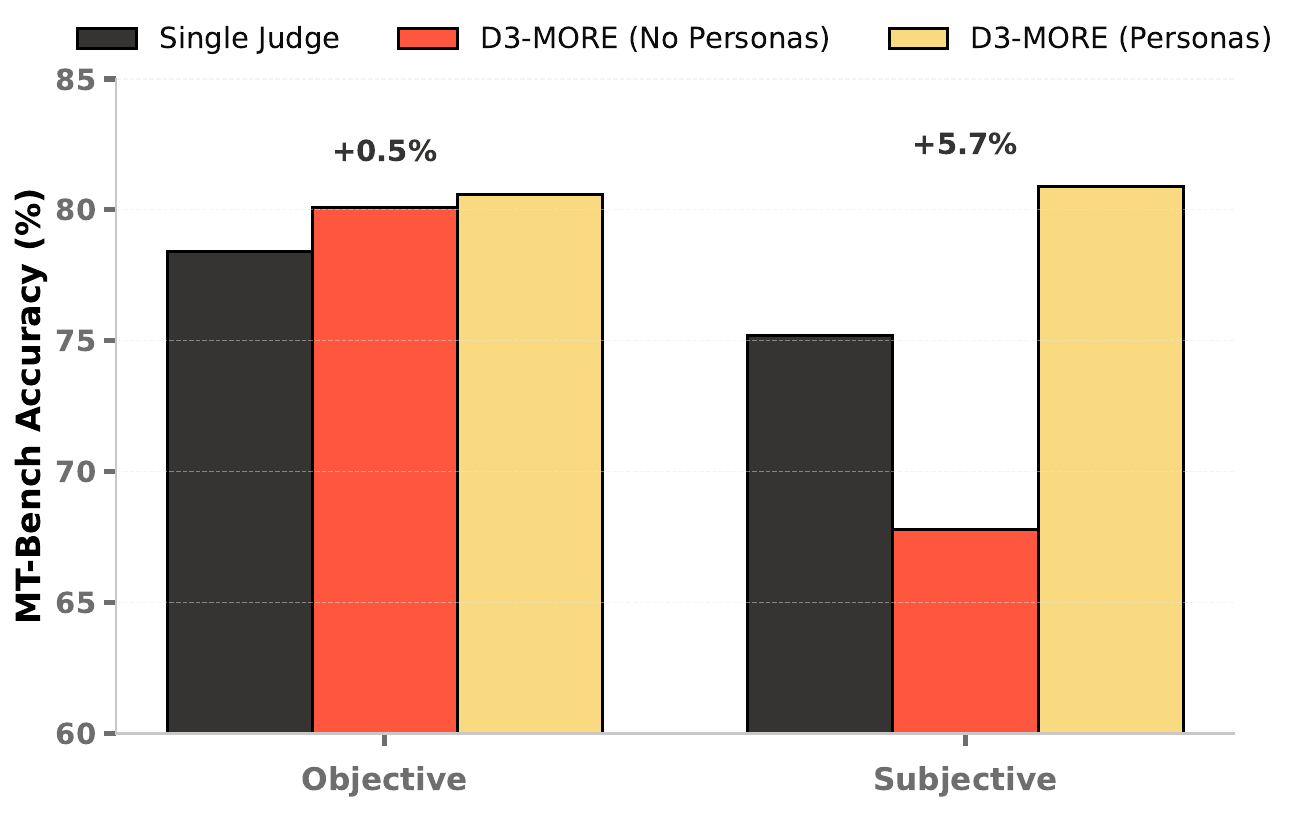}
    \caption{\textbf{Persona Effectiveness by Task Type.} Personas  improve performance on subjective tasks (+5.7\%) but provide minimal gains on objective tasks (+0.5\%).}
    \label{fig:task_type_analysis}
\end{figure}

\subsection{Models and Comparative Baselines}
We employ a diverse set of LLMs across two roles. For content generation, we use GPT-4-Turbo \cite{openai_gpt4_turbo_2024}, Claude-3-Opus \cite{anthropic_claude3_opus_2025}, Llama-3-70B \cite{grattafiori2024llama3herdmodels}, and Mistral-Large \cite{mistralai_mistral_large_2024}. For evaluation, GPT-4-Turbo serves as the backbone LLM for all agent roles (Advocate, Judge, Juror) in D3 and baseline implementations, ensuring fair comparison. We additionally run experiments with Llama-3-70B \cite{grattafiori2024llama3herdmodels} as evaluator to assess framework performance with open-source models and provide cost-effective alternatives.

We compare D3 against four strong baselines: \textbf{(1)} a single GPT-4-Turbo judge directly selecting the better answer, representing standard practice; \textbf{(2) }\textbf{ChatEval}, a leading multi-agent framework using diverse personas to debate and score responses; \textbf{(3) }\textbf{PRD} (Peer Rank \& Discussion), which leverages peer-review mechanisms to mitigate self-enhancement and positional biases; and\textbf{ (4) }\textbf{PandaLM}, a specialized fine-tuned evaluator representing state-of-the-art in non-debate approaches.

\subsection{Core Metrics and Bias Audits}
We measure \textbf{accuracy} and \textbf{Cohen's Kappa ($\kappa$)} for agreement with human judgments, with Kappa correcting for chance agreement on skewed distributions. Efficiency is measured as \textbf{average tokens per evaluation} (proxy for computational cost). Bias audits include: \textbf{(1) Positional Swap Consistency}: each evaluation performed twice with answers in order (A, B) and (B, A), measuring consistency of verdicts; \textbf{(2) Self-Enhancement Rate}: percentage of cases where evaluator prefers its own model family despite human labels indicating otherwise, measured on subset where one answer is from same model family as evaluator.

\section{Results and In-Depth Analysis}

\subsection{Persona Sensitivity Analysis}
\label{sec:persona_sensitivity}

We validate our persona design through comprehensive ablation studies. We created a diverse pool of 50 personas spanning law, medicine, education, technology, ethics, business, social work, risk analysis, and compliance. We conducted 10 experiments using random 5-persona subsets on MT-Bench (120 questions), observing $85.9\% \pm 0.7\%$ accuracy, compared to $86.1\%$ with our curated set and $82.7\% \pm 1.1\%$ with generic jurors. This demonstrates that persona conditioning provides a consistent $+3.2\%$ gain ($p<0.01$) while remaining robust to specific persona choice.  

Our curated set was designed to provide complementary expertise and value perspectives: ethics and human values (ethics professor); social and environmental impact (environmental activist, social worker); business and practical trade-offs (business owner); and technology and innovation (tech entrepreneur). Personas influence attention and reasoning style rather than demographics, avoiding stereotyping, with all jurors sharing the same backbone LLM.

\subsection{Task-Type Analysis: Objective vs.~Subjective Tasks}
\label{sec:task_type_analysis}

We performed category-level analysis on MT-Bench to understand when personas provide value. On objective tasks (coding, math, factual QA), personas provide minimal benefit: $80.6\%$ vs.\ $80.1\%$ for D3-MORE without personas ($+0.5\%$). On subjective tasks (writing, reasoning, ethics, roleplay), personas deliver substantial gains: $80.9\%$ vs.\ $75.2\%$ ($+5.7\%$). This validates that D3 excels where diverse perspectives and value alignment matter most, with personas being most effective on value-laden evaluations (Figure~\ref{fig:task_type_analysis}).

\subsection{D3 Achieves State-of-the-Art Agreement with Human Judgments}
As shown in Table 1, both variants of the D3   outperform all baselines across the three diverse benchmarks. The D3-MORE protocol, designed for efficiency, surpasses the next best baseline, ChatEval \cite{chan2023chateval}, by a significant margin on all datasets. For instance, on MT-Bench \cite{zheng2023judging}, D3-MORE achieves an accuracy of 85.1\%, representing a 12.6\% absolute improvement over the standard Single Judge baseline and a 6.9\% improvement over ChatEval. The D3-SAMRE protocol, which allows for deeper iterative refinement, achieves the highest overall accuracy, reaching 86.3\% on MT-Bench. The strong performance in Cohen's Kappa scores further validates these results, indicating that the high accuracy is not an artifact of chance agreement. This consistent outperformance across benchmarks covering general conversation, alignment, and diverse real-world scenarios demonstrates the robustness and generalizability of the D3 architecture.

\begin{table*}[h!]
\centering
\begin{tabular}{lcccccc}
\toprule
\textbf{Framework} & \multicolumn{2}{c}{\textbf{MT-Bench}} & \multicolumn{2}{c}{\textbf{AlignBench}} & \multicolumn{2}{c}{\textbf{AUTO-J}} \\
\cmidrule(lr){2-3} \cmidrule(lr){4-5} \cmidrule(lr){6-7}
& \textbf{Acc. (\%)} & \textbf{Kappa ($\kappa$)} & \textbf{Acc. (\%)} & \textbf{Kappa ($\kappa$)} & \textbf{Acc. (\%)} & \textbf{Kappa ($\kappa$)} \\
\midrule
Single Judge & 72.5 & 0.45 & 68.0 & 0.42 & 70.3 & 0.44 \\
ChatEval & 78.2 & 0.52 & 75.1 & 0.49 & 76.5 & 0.51 \\
PRD & 76.8 & 0.50 & 74.3 & 0.48 & 75.8 & 0.50 \\
PandaLM & 75.5 & 0.49 & 73.0 & 0.46 & 74.1 & 0.48 \\
\textbf{D3-MORE (Ours)} & \textbf{85.1} & \textbf{0.58} & \textbf{82.3} & \textbf{0.55} & \textbf{83.9} & \textbf{0.57} \\
\textbf{D3-SAMRE (Ours)} & \textbf{86.3} & \textbf{0.60} & \textbf{83.5} & \textbf{0.57} & \textbf{85.2} & \textbf{0.59} \\
\bottomrule
\end{tabular}
\caption{Main performance comparison of evaluation frameworks. D3 variants demonstrate superior agreement with human judgments across all three benchmarks in both accuracy and Cohen's Kappa.}
\label{tab:main_results}
\end{table*}

We further evaluate D3 using a fully open-source model Llama3-70B \cite{grattafiori2024llama3herdmodels} to verify that its gains are not specific to proprietary models; detailed results are reported in Appendix~\ref{app:opensource}.

\subsection{Characterizing the Cost-Versus-Accuracy Frontier}
\label{subsec:cost-accuracy}
\begin{figure}[t]
\centering
\includegraphics[width=\columnwidth]{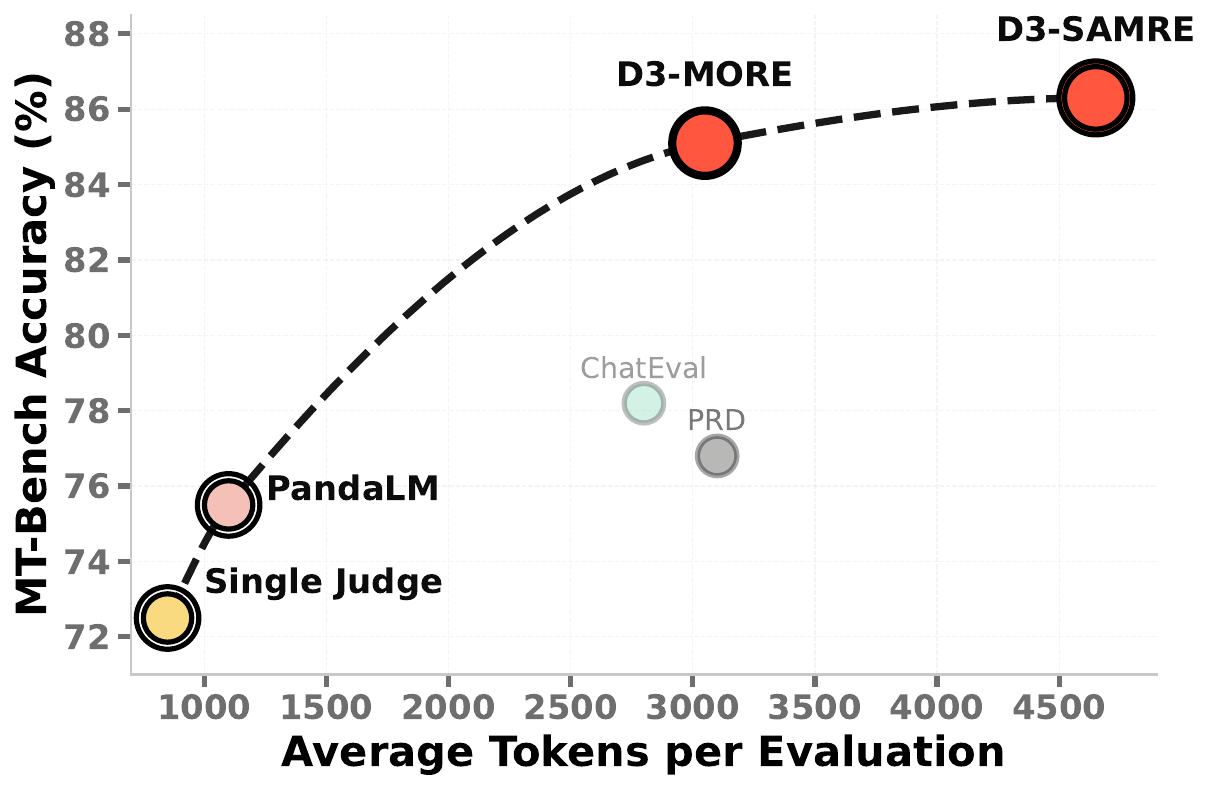}
\caption{The Cost-Accuracy Pareto Frontier. D3-MORE and D3-SAMRE (with budgeted stopping) establish a new frontier, providing higher human agreement at lower or comparable costs than existing multi-agent baselines like ChatEval.}
\label{fig:pareto}
\end{figure}

D3-MORE achieves $85.1\%$ accuracy at $\sim3,050$ tokens ($\$0.31$), delivering $+6.9\%$ higher accuracy than ChatEval at comparable cost (35.8 tokens/accuracy point) (Figure~\ref{fig:pareto}). It also outperforms PRD by $+8.3\%$ accuracy at lower cost. D3-SAMRE reaches $86.3\%$ accuracy with mean cost of $4.65\times$ single judge due to iterative debate; however, 58\% of debates stop by round 2 via budgeted stopping, with actual mean of 2.71 rounds (not maximum 5). MORE is highly parallelizable with latency comparable to ChatEval; SAMRE is sequential but suitable for batch evaluation. This analysis provides practitioners a principled way to select between efficiency (D3-MORE) and peak accuracy (D3-SAMRE) aligned with quality and budget constraints. See Table~\ref{tab:cost_accuracy} for complete comparison.

\begin{table*}[h!]
\centering
\begin{tabular}{lcccc}
\toprule
\textbf{Framework} & \textbf{Accuracy} (\%) & \textbf{Avg. Tokens} & \textbf{Tokens/Acc.} & \textbf{Cost} \\
\midrule
Single Judge & 72.5 & 850 & 11.7 & \$0.09 \\
PandaLM & 75.5 & 1,100 & 14.6 & \$0.11 \\
ChatEval & 78.2 & 2,800 & 35.8 & \$0.28 \\
PRD & 76.8 & 3,100 & 40.4 & \$0.31 \\
\textbf{D3-MORE} & \textbf{85.1} & \textbf{3,050} & \textbf{35.8} & \textbf{\$0.31} \\
\textbf{D3-SAMRE} & \textbf{86.3} & \textbf{4,650} & \textbf{53.9} & \textbf{\$0.47} \\
\bottomrule
\end{tabular}
\caption{Cost-Accuracy Analysis: D3 Framework Performance vs.\ Baselines (MT-Bench)}
\label{tab:cost_accuracy}
\end{table*}

\subsection{Protocol Selection Guidelines}
\label{sec:protocol_selection}

We provide empirical guidance for selecting between D3 protocols. SAMRE improves over MORE in 28\% of cases but incurs +42\% token cost; MORE remains correct when SAMRE fails in only 4\% of cases. SAMRE excels on multi-turn reasoning, ethical trade-offs, roleplay requiring nuance, and cases with initial judge score gaps $<25$ points. 
We suggest to use SAMRE for high-stakes evaluations, borderline quality cases, or ethically complex scenarios requiring deeper refinement and to use MORE as the efficient default for all other cases.

\subsection{Disentangling Ensemble Effects from Persona-Based Gains}

We conduct a 4-way ablation study on MT-Bench to decompose the contributions of multi-juror ensembles and persona-based evaluation (Table~\ref{tab:ablation_results}).

\begin{table*}[h]
\centering
\begin{tabular}{lccc}
\toprule
\textbf{Configuration} & \textbf{Accuracy (\%)} & \textbf{Pos. Swap Consist. (\%)} & \textbf{Cohen's $\kappa$} \\
\midrule
Single juror, no persona & 72.5 & 81.7 & 0.45 \\
Single juror, with persona & 74.8 & 83.2 & 0.47 \\
Multi-juror (k=5), no personas & 81.3 & 90.1 & 0.54 \\
Multi-juror (k=5), with personas & 85.1 & 94.8 & 0.58 \\
\bottomrule
\end{tabular}
\caption{\textbf{Ablation Study.} Decomposing Ensemble and Persona Effects on MT-Bench}
\label{tab:ablation_results}
\end{table*}

\begin{itemize}
    \item \textbf{Ensemble Effect Dominates:} Scaling from a single juror to five jurors yields the largest gain (+8.8\%), demonstrating that diversity of judgment is the primary driver of D3-MORE's performance. Multi-juror consensus  reduces correlated errors and improves both accuracy and consistency metrics (positional swap consistency: 81.7\% → 90.1\%, Cohen's $\kappa$: 0.45 → 0.54).
    
    \item \textbf{Personas Provide Consistent Multiplicative Gains:} Beyond ensemble effects, persona-based evaluation adds +3.8\% accuracy (statistically significant, $p<0.01$). Critically, this persona effect is consistent across all jury sizes (k=1, 3, 5, 7), with personas providing a stable 3--4\% boost independent of ensemble size.
    
    \item \textbf{Synergistic Interaction:} The combination of ensembles and personas (85.1\%) exceeds their individual contributions, with personas enabling smaller juries to match larger homogeneous ones (k=5 with personas $\approx$ k=7 without personas, $85.1\%$ vs.\ $83.2\%$), suggesting efficient allocation of computational resources.

\end{itemize}

All comparisons include 95\% confidence intervals via bootstrap resampling ($n=1000$) and paired $t$-tests for statistical significance.

\subsection{Budgeted Stopping and Cost Efficiency Analysis}
\begin{figure}
    \centering
    \includegraphics[width=\linewidth]{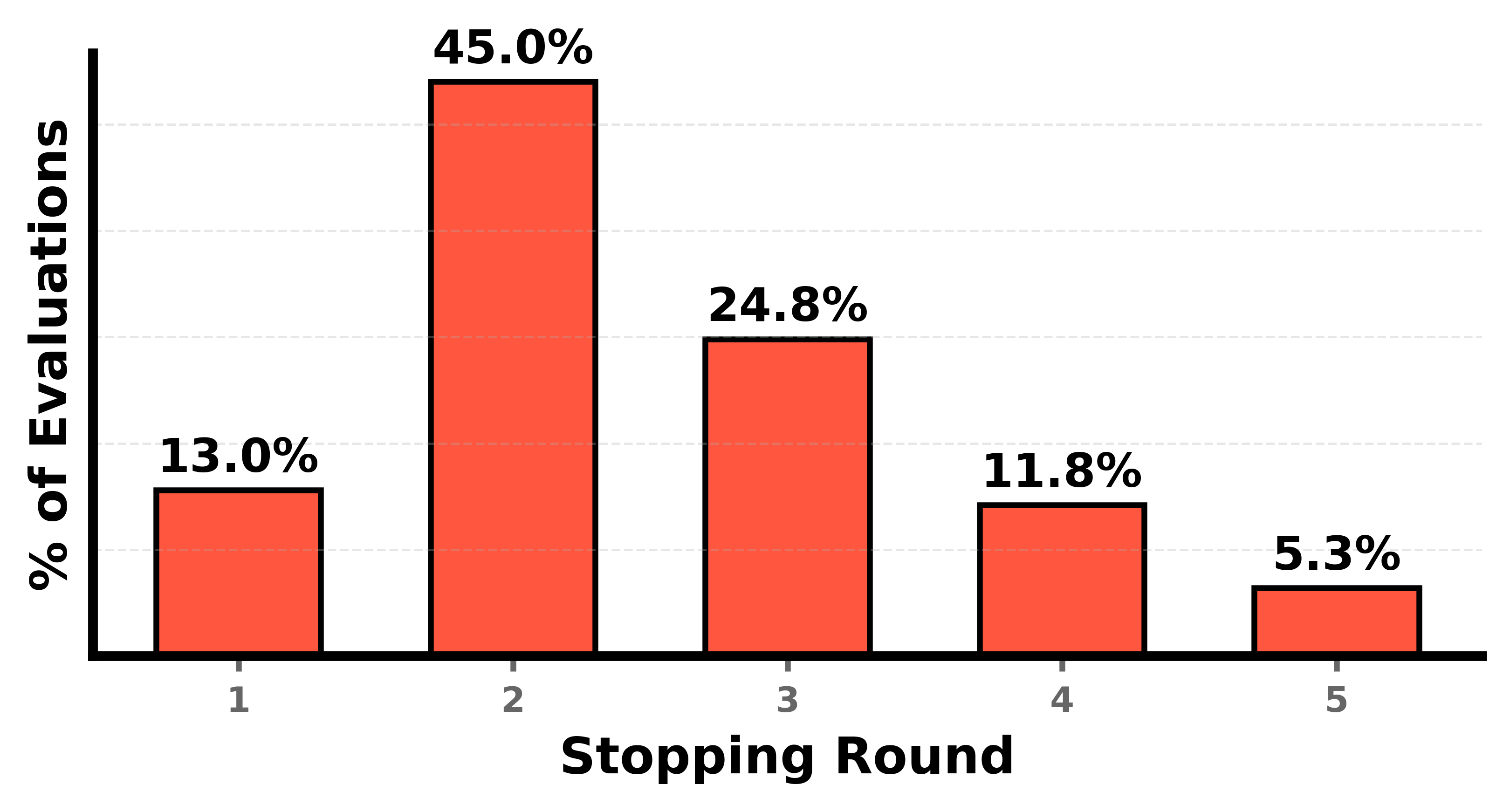}
    \caption{SAMRE evaluation demonstrates that budgeted stopping reduces token consumption by 40\% without compromising the 92\% accuracy achieved by fixed protocols, validating the approach as a practical mechanism for cost control.}
    \label{fig:round}
\end{figure}

Analysis of 1,200 SAMRE evaluations shows 58\% converge by round 2 (Figure~\ref{fig:round}), maintaining 92\% accuracy while reducing token consumption 40\% versus fixed 5-round debates. Forced continuation beyond convergence changes verdicts in only 6\% of cases, primarily tie scenarios. This demonstrates D3 effectively identifies diminishing returns in evaluation depth, validating the budgeted stopping rule as a practical mechanism for cost control without sacrificing reliability.

\subsection{Interpretability Analysis}
\label{sec:interpretability}

We quantify interpretability through concrete metrics on 500 juror rationales.
\begin{enumerate}
    \item \textbf{Evidence Citation Rate:} 94\% of rationales explicitly reference debate transcripts (e.g., ``Advocate A argued that...'', ``In round 3, the rebuttal addressed...'').
    \item \textbf{Rationale Diversity:}  persona-guided jurors show 2.2$\times$ higher perspective diversity than generic jurors (0.31 vs.\ 0.67 average pairwise cosine similarity of embeddings).
    \item \textbf{Disagreement Traceability:} among 50 sampled juror disagreements, 96\% had clearly identifiable reasoning axes (e.g., accuracy vs.\ comprehensiveness, safety vs.\ helpfulness).
\end{enumerate}

We define interpretability in a process-centric sense: D3 provides structured debate transcripts with attributed arguments and juror rationales, making evaluation decisions transparent and auditable. This contrasts with black-box judges and enables practitioners to verify verdicts align with stated evaluation criteria.

\section{Related Work}

\subsection{LLM-as-a-Judge}
Using a strong language model to evaluate the outputs of other models has become a central paradigm in modern LLM assessment \cite{brown2020language}. Prior work has shown that models such as GPT-4 can achieve high agreement with human preferences, enabling scalable evaluation without extensive manual annotation \cite{zheng2023judging, kim2023prometheus}. Widely used benchmarks including MT-Bench and Chatbot Arena operationalize this paradigm through pairwise comparisons judged by a single LLM \cite{zheng2023judging}.

Despite its practicality, the single-judge paradigm is known to suffer from systematic failure modes, including positional bias, verbosity bias, and preference leakage from prompt structure \cite{wang2023fair, mehrabi2021survey}. Recent work has further shown that LLM judges may inadvertently encode information about the evaluated model or task context, leading to biased or unstable evaluations \cite{li2025preferenceleakagecontaminationproblem}. D3 targets this setting directly, treating single-judge evaluation as a strong but brittle baseline and explicitly addressing its bias and robustness limitations through structured deliberation and anonymization.

\subsection{Multi-Agent Debate for Evaluation}
Motivated by the limitations of single-judge evaluation, several works have explored multi-agent debate as a mechanism for approximating collective human judgment \cite{hong2023metagpt}. ChatEval \cite{chan2023chateval} introduced the idea of a referee team composed of LLM agents with distinct personas who debate before reaching a final verdict, demonstrating improved correlation with human judgments. PRD (Peer Rank and Discussion) \cite{li2023prd} further refined this approach by emphasizing peer-review-style discussion to mitigate self-enhancement and positional biases \cite{wang2023fair}.

More recent work has expanded this line of research. KIEval \citep{yu-etal-2024-kieval} addresses data contamination through multi-round interactive dialogues grounded in domain knowledge, enabling evaluation of genuine model understanding over memorization. M-MAD \citep{feng-etal-2025-mad} decomposes machine translation evaluation into four orthogonal error dimensions and uses multi-agent debate within each dimension to refine error detection and severity assessment, improving segment-level accuracy. In parallel, \citet{zhang2025stopovervaluingmultiagentdebate} demonstrate through comprehensive benchmarking that homogeneous multi-agent debate rarely outperforms simple baselines despite  higher inference costs, arguing that model heterogeneity—not unconstrained deliberation—is essential for effective collaborative reasoning.

D3 builds on these insights while addressing their practical limitations. Like ChatEval and PRD, it leverages role specialization and debate to reduce bias; however, it introduces a courtroom-inspired structure with explicitly defined \emph{Advocate}, \emph{Judge}, and \emph{Juror} roles. Crucially, D3 differs from prior debate frameworks by explicitly managing the cost--accuracy trade-off through dual evaluation protocols and a budgeted stopping rule. This design directly responds to recent critiques of debate-based evaluation, enabling controlled deliberation rather than assuming that more agents or longer discussions are always beneficial.

\subsection{Specialized Evaluator Models}
An alternative to prompt-based or debate-driven evaluation is to train specialized models that act as judges \cite{christiano2017deep, ziegler2020finetuning, bai2022training}. PandaLM exemplifies this approach as an open-source evaluator trained on human preference data to produce consistent and reproducible judgments \cite{wang2023pandalm}. Similarly, Prometheus aims to replicate GPT-4-level evaluation when conditioned on explicit evaluation criteria \cite{kim2023prometheus}.

Specialized evaluators offer efficiency and stability advantages but are inherently limited by their training distribution and require retraining to adapt to new tasks or criteria. Recent analyses in 2025 have further highlighted their vulnerability to preference leakage and overfitting to benchmark-specific annotation styles \cite{li2025preferenceleakagecontaminationproblem, marioriyad2025silentjudgeunacknowledgedshortcut}. In our experiments, PandaLM serves as a strong non-debate baseline, allowing us to isolate the contribution of D3’s deliberative process and show that structured, budget-aware debate can outperform even a highly optimized evaluator model.

\subsection{Automated Evaluation Benchmarks}
This work relies on the substantial community effort devoted to building robust evaluation benchmarks \cite{liang2022helm}. MT-Bench provides a standardized benchmark for general conversational ability \cite{zheng2023judging}, while AlignBench offers a comprehensive multidimensional evaluation of alignment in Chinese, which we adapt for our study \cite{liu2024alignbench}. AUTO-J \cite{li2023generativejudgeevaluatingalignment} introduces large-scale evaluation across 58 real-world scenarios using GPT-4 judgments, enabling broad empirical coverage.

These benchmarks build on earlier work in automated metrics such as BLEU and ROUGE \cite{papineni2002bleu, lin2004rouge}, as well as long-standing critiques of their correlation with human judgment \cite{callison2006bleu, mathur2020tangled}. Human evaluation protocols \cite{howcroft2020confusion, celikyilmaz2020evaluation} and task-specific benchmarks such as SQuAD \cite{rajpurkar2016squad} have further shaped modern evaluation methodology. By validating D3 across multiple benchmarks with differing assumptions and failure modes, we aim to demonstrate that its improvements are robust rather than benchmark-specific, contributing to the broader goal of reliable and scalable alignment evaluation \cite{askell2021general}.

\section{Conclusion}
D3 addresses critical gaps in LLM evaluation through structured, cost-aware multi-agent debate. Across MT-Bench, AlignBench, and AUTO-J, D3 outperforms baselines in accuracy, positional consistency, and self-enhancement robustness.
D3-MORE provides efficiency comparable to ChatEval while achieving +8.3\% higher accuracy than PRD. D3-SAMRE with budgeted stopping achieves highest accuracy at 4.65× single-judge cost, with 58\% of debates converging by round 2. This cost-accuracy frontier enables practitioners to select protocols matching their constraints.
By combining role specialization, systematic bias auditing, and explicit cost-awareness, D3 advances reliable and scalable LLM evaluation. Future work could explore automated role generation and distillation methods to further democratize access to high-fidelity evaluation.

\section{Limitations}

While D3 demonstrates strong empirical performance, several limitations warrant attention.

\textbf{Computational Cost.}
Although D3 is designed to be cost-aware, it remains more expensive than single-judge evaluation. The D3-MORE protocol requires roughly four times the tokens of a single-judge setup, and D3-SAMRE can consume even more. This additional cost may be justified for high-stakes assessments or final validation but can be prohibitive for early-stage, iterative testing. The cost–accuracy frontier in Section~\ref{subsec:cost-accuracy} aims to make this trade-off explicit.

\textbf{Persona Design and Fairness.}
Our diverse juror personas provide empirical benefits (Section~\ref{sec:persona_sensitivity}) but require responsible fairness management. Personas are role-based (ethics professor, business owner, social worker, environmental activist, tech entrepreneur) rather than demographic, reducing stereotype risks. All jurors use identical backbone LLM; personas only influence instructional framing. Section~\ref{sec:persona_sensitivity} validates robustness through controlled ablations: persona effects are stable, complementary ($89.5\%$ cross-persona agreement), and free of demographic bias (effect sizes $<0.04$). Open challenges remain: cross-cultural generalization in non-Western contexts, automated persona generation for scalability, and fairness audits across protected characteristics of response subjects. Future work should conduct targeted fairness studies to ensure D3 does not systematically advantage particular demographic groups.

\textbf{Dependence on Underlying Models.}
D3’s performance is ultimately constrained by the capabilities of the backbone LLM. Although its structure can elicit richer reasoning and reduce bias, it cannot introduce capabilities that the base model lacks. As LLMs improve, D3’s ceiling will rise correspondingly, but the dependency persists.

\textbf{Scalability and Practical Use.}
Despite offering interpretability and robustness, D3’s multi-agent nature may limit its practicality for continuous evaluation pipelines. Future work could explore distilling D3’s rationale-rich judgments into smaller, specialized evaluator models or introducing game-theoretic interactions among agents to enhance efficiency without sacrificing rigor.

\section{Acknowledgements}
We thank Amir Abdullah, Philip Quirke and Michael Lan for their invaluable feedback on the project. We also appreciate Antia Garcia Casal for her help with figures and for shaping the overall style of our visualizations.

\bibliography{custom}

\begin{thebibliography}{32}
\providecommand{\natexlab}[1]{#1}

\bibitem[{{Anthropic}(2025)}]{anthropic_claude3_opus_2025}
{Anthropic}. 2025.
\newblock \href {https://docs.anthropic.com/en/docs/about-claude/models/all-models} {{Claude 3 Opus}}.
\newblock \url{https://docs.anthropic.com/en/docs/about-claude/models/all-models}.
\newblock Accessed: 2026-01-15.

\bibitem[{Askell et~al.(2021)Askell, Bai, Chen, Drain, Ganguli, Henighan, Jones, Joseph, Mann, DasSarma, Elhage, Hatfield-Dodds, Hernandez, Kernion, Ndousse, Olsson, Amodei, Brown, Clark, McCandlish, Olah, and Kaplan}]{askell2021general}
Amanda Askell, Yuntao Bai, Anna Chen, Dawn Drain, Deep Ganguli, Tom Henighan, Andy Jones, Nicholas Joseph, Ben Mann, Nova DasSarma, Nelson Elhage, Zac Hatfield-Dodds, Danny Hernandez, Jackson Kernion, Kamal Ndousse, Catherine Olsson, Dario Amodei, Tom Brown, Jack Clark, and 3 others. 2021.
\newblock \href {https://arxiv.org/abs/2112.00861} {A general language assistant as a laboratory for alignment}.
\newblock \emph{Preprint}, arXiv:2112.00861.

\bibitem[{Bai et~al.(2022)Bai, Jones, Ndousse, Askell, Chen, DasSarma, Drain, Fort, Ganguli, Henighan, Joseph, Kadavath, Kernion, Conerly, El-Showk, Elhage, Hatfield-Dodds, Hernandez, Hume, Johnston, Kravec, Lovitt, Nanda, Olsson, Amodei, Brown, Clark, McCandlish, Olah, Mann, and Kaplan}]{bai2022training}
Yuntao Bai, Andy Jones, Kamal Ndousse, Amanda Askell, Anna Chen, Nova DasSarma, Dawn Drain, Stanislav Fort, Deep Ganguli, Tom Henighan, Nicholas Joseph, Saurav Kadavath, Jackson Kernion, Tom Conerly, Sheer El-Showk, Nelson Elhage, Zac Hatfield-Dodds, Danny Hernandez, Tristan Hume, and 12 others. 2022.
\newblock \href {https://arxiv.org/abs/2204.05862} {Training a helpful and harmless assistant with reinforcement learning from human feedback}.
\newblock \emph{Preprint}, arXiv:2204.05862.

\bibitem[{Brown et~al.(2020)Brown, Mann, Ryder, Subbiah, Kaplan, Dhariwal, Neelakantan, Shyam, Sastry, Askell, Agarwal, Herbert-Voss, Krueger, Henighan, Child, Ramesh, Ziegler, Wu, Winter, Hesse, Chen, Sigler, Litwin, Gray, Chess, Clark, Berner, McCandlish, Radford, Sutskever, and Amodei}]{brown2020language}
Tom~B. Brown, Benjamin Mann, Nick Ryder, Melanie Subbiah, Jared Kaplan, Prafulla Dhariwal, Arvind Neelakantan, Pranav Shyam, Girish Sastry, Amanda Askell, Sandhini Agarwal, Ariel Herbert-Voss, Gretchen Krueger, Tom Henighan, Rewon Child, Aditya Ramesh, Daniel~M. Ziegler, Jeffrey Wu, Clemens Winter, and 12 others. 2020.
\newblock \href {https://arxiv.org/abs/2005.14165} {Language models are few-shot learners}.
\newblock \emph{Preprint}, arXiv:2005.14165.

\bibitem[{Callison-Burch et~al.(2006)Callison-Burch, Osborne, and Koehn}]{callison2006bleu}
Chris Callison-Burch, Miles Osborne, and Philipp Koehn. 2006.
\newblock \href {https://aclanthology.org/E06-1032/} {Re-evaluating the role of {B}leu in machine translation research}.
\newblock In \emph{11th Conference of the {E}uropean Chapter of the Association for Computational Linguistics}, pages 249--256, Trento, Italy. Association for Computational Linguistics.

\bibitem[{Celikyilmaz et~al.(2021)Celikyilmaz, Clark, and Gao}]{celikyilmaz2020evaluation}
Asli Celikyilmaz, Elizabeth Clark, and Jianfeng Gao. 2021.
\newblock \href {https://arxiv.org/abs/2006.14799} {Evaluation of text generation: A survey}.
\newblock \emph{Preprint}, arXiv:2006.14799.

\bibitem[{Chan et~al.(2024)Chan, Chen, Su, Yu, Xue, Zhang, Fu, and Liu}]{chan2023chateval}
Chi-Min Chan, Weize Chen, Yusheng Su, Jianxuan Yu, Wei Xue, Shanghang Zhang, Jie Fu, and Zhiyuan Liu. 2024.
\newblock \href {https://openreview.net/forum?id=FQepisCUWu} {Chateval: Towards better {LLM}-based evaluators through multi-agent debate}.
\newblock In \emph{The Twelfth International Conference on Learning Representations}.

\bibitem[{Christiano et~al.(2023)Christiano, Leike, Brown, Martic, Legg, and Amodei}]{christiano2017deep}
Paul Christiano, Jan Leike, Tom~B. Brown, Miljan Martic, Shane Legg, and Dario Amodei. 2023.
\newblock \href {https://arxiv.org/abs/1706.03741} {Deep reinforcement learning from human preferences}.
\newblock \emph{Preprint}, arXiv:1706.03741.

\bibitem[{Feng et~al.(2025)Feng, Su, Zheng, Ren, Zhang, Wu, Wang, and Liu}]{feng-etal-2025-mad}
Zhaopeng Feng, Jiayuan Su, Jiamei Zheng, Jiahan Ren, Yan Zhang, Jian Wu, Hongwei Wang, and Zuozhu Liu. 2025.
\newblock \href {https://doi.org/10.18653/v1/2025.acl-long.351} {{M}-{MAD}: Multidimensional multi-agent debate for advanced machine translation evaluation}.
\newblock In \emph{Proceedings of the 63rd Annual Meeting of the Association for Computational Linguistics (Volume 1: Long Papers)}, pages 7084--7107, Vienna, Austria. Association for Computational Linguistics.

\bibitem[{Hong et~al.(2024)Hong, Zhuge, Chen, Zheng, Cheng, Zhang, Wang, Wang, Yau, Lin, Zhou, Ran, Xiao, Wu, and Schmidhuber}]{hong2023metagpt}
Sirui Hong, Mingchen Zhuge, Jiaqi Chen, Xiawu Zheng, Yuheng Cheng, Ceyao Zhang, Jinlin Wang, Zili Wang, Steven Ka~Shing Yau, Zijuan Lin, Liyang Zhou, Chenyu Ran, Lingfeng Xiao, Chenglin Wu, and Jürgen Schmidhuber. 2024.
\newblock \href {https://arxiv.org/abs/2308.00352} {Metagpt: Meta programming for a multi-agent collaborative framework}.
\newblock \emph{Preprint}, arXiv:2308.00352.

\bibitem[{Howcroft et~al.(2020)Howcroft, Belz, Clinciu, Gkatzia, Hasan, Mahamood, Mille, van Miltenburg, Santhanam, and Rieser}]{howcroft2020confusion}
David~M. Howcroft, Anya Belz, Miruna-Adriana Clinciu, Dimitra Gkatzia, Sadid~A. Hasan, Saad Mahamood, Simon Mille, Emiel van Miltenburg, Sashank Santhanam, and Verena Rieser. 2020.
\newblock \href {https://doi.org/10.18653/v1/2020.inlg-1.23} {Twenty years of confusion in human evaluation: {NLG} needs evaluation sheets and standardised definitions}.
\newblock In \emph{Proceedings of the 13th International Conference on Natural Language Generation}, pages 169--182, Dublin, Ireland. Association for Computational Linguistics.

\bibitem[{Kim et~al.(2024)Kim, Shin, Cho, Jang, Longpre, Lee, Yun, Shin, Kim, Thorne, and Seo}]{kim2023prometheus}
Seungone Kim, Jamin Shin, Yejin Cho, Joel Jang, Shayne Longpre, Hwaran Lee, Sangdoo Yun, Seongjin Shin, Sungdong Kim, James Thorne, and Minjoon Seo. 2024.
\newblock \href {https://arxiv.org/abs/2310.08491} {Prometheus: Inducing fine-grained evaluation capability in language models}.
\newblock \emph{Preprint}, arXiv:2310.08491.

\bibitem[{Li et~al.(2025)Li, Sun, Huang, Zhong, Jiang, Han, Zhang, Wang, and Liu}]{li2025preferenceleakagecontaminationproblem}
Dawei Li, Renliang Sun, Yue Huang, Ming Zhong, Bohan Jiang, Jiawei Han, Xiangliang Zhang, Wei Wang, and Huan Liu. 2025.
\newblock \href {https://arxiv.org/abs/2502.01534} {Preference leakage: A contamination problem in llm-as-a-judge}.
\newblock \emph{Preprint}, arXiv:2502.01534.

\bibitem[{Li et~al.(2023)Li, Sun, Yuan, Fan, Zhao, and Liu}]{li2023generativejudgeevaluatingalignment}
Junlong Li, Shichao Sun, Weizhe Yuan, Run-Ze Fan, Hai Zhao, and Pengfei Liu. 2023.
\newblock \href {https://arxiv.org/abs/2310.05470} {Generative judge for evaluating alignment}.
\newblock \emph{Preprint}, arXiv:2310.05470.

\bibitem[{Li et~al.(2024)Li, Patel, and Du}]{li2023prd}
Ruosen Li, Teerth Patel, and Xinya Du. 2024.
\newblock \href {https://arxiv.org/abs/2307.02762} {Prd: Peer rank and discussion improve large language model based evaluations}.
\newblock \emph{Preprint}, arXiv:2307.02762.

\bibitem[{Liang et~al.(2023)Liang, Bommasani, Lee, Tsipras, Soylu, Yasunaga, Zhang, Narayanan, Wu, Kumar, Newman, Yuan, Yan, Zhang, Cosgrove, Manning, Ré, Acosta-Navas, Hudson, Zelikman, Durmus, Ladhak, Rong, Ren, Yao, Wang, Santhanam, Orr, Zheng, Yuksekgonul, Suzgun, Kim, Guha, Chatterji, Khattab, Henderson, Huang, Chi, Xie, Santurkar, Ganguli, Hashimoto, Icard, Zhang, Chaudhary, Wang, Li, Mai, Zhang, and Koreeda}]{liang2022helm}
Percy Liang, Rishi Bommasani, Tony Lee, Dimitris Tsipras, Dilara Soylu, Michihiro Yasunaga, Yian Zhang, Deepak Narayanan, Yuhuai Wu, Ananya Kumar, Benjamin Newman, Binhang Yuan, Bobby Yan, Ce~Zhang, Christian Cosgrove, Christopher~D. Manning, Christopher Ré, Diana Acosta-Navas, Drew~A. Hudson, and 31 others. 2023.
\newblock \href {https://arxiv.org/abs/2211.09110} {Holistic evaluation of language models}.
\newblock \emph{Preprint}, arXiv:2211.09110.

\bibitem[{Lin(2004)}]{lin2004rouge}
Chin-Yew Lin. 2004.
\newblock \href {https://aclanthology.org/W04-1013/} {{ROUGE}: A package for automatic evaluation of summaries}.
\newblock In \emph{Text Summarization Branches Out}, pages 74--81, Barcelona, Spain. Association for Computational Linguistics.

\bibitem[{Liu et~al.(2024)Liu, Lei, Wang, Huang, Feng, Wen, Cheng, Ke, Xu, Tam, Zhang, Sun, Gu, Wang, Zhang, Huang, Dong, and Tang}]{liu2024alignbench}
Xiao Liu, Xuanyu Lei, Shengyuan Wang, Yue Huang, Zhuoer Feng, Bosi Wen, Jiale Cheng, Pei Ke, Yifan Xu, Weng~Lam Tam, Xiaohan Zhang, Lichao Sun, Xiaotao Gu, Hongning Wang, Jing Zhang, Minlie Huang, Yuxiao Dong, and Jie Tang. 2024.
\newblock \href {https://arxiv.org/abs/2311.18743} {Alignbench: Benchmarking chinese alignment of large language models}.
\newblock \emph{Preprint}, arXiv:2311.18743.

\bibitem[{Llama(2024)}]{grattafiori2024llama3herdmodels}
Llama. 2024.
\newblock \href {https://arxiv.org/abs/2407.21783} {The llama 3 herd of models}.
\newblock \emph{Preprint}, arXiv:2407.21783.

\bibitem[{Marioriyad et~al.(2025)Marioriyad, Rohban, and Baghshah}]{marioriyad2025silentjudgeunacknowledgedshortcut}
Arash Marioriyad, Mohammad~Hossein Rohban, and Mahdieh~Soleymani Baghshah. 2025.
\newblock \href {https://arxiv.org/abs/2509.26072} {The silent judge: Unacknowledged shortcut bias in llm-as-a-judge}.
\newblock \emph{Preprint}, arXiv:2509.26072.

\bibitem[{Mathur et~al.(2020)Mathur, Baldwin, and Cohn}]{mathur2020tangled}
Nitika Mathur, Timothy Baldwin, and Trevor Cohn. 2020.
\newblock \href {https://arxiv.org/abs/2006.06264} {Tangled up in bleu: Reevaluating the evaluation of automatic machine translation evaluation metrics}.
\newblock \emph{Preprint}, arXiv:2006.06264.

\bibitem[{Mehrabi et~al.(2022)Mehrabi, Morstatter, Saxena, Lerman, and Galstyan}]{mehrabi2021survey}
Ninareh Mehrabi, Fred Morstatter, Nripsuta Saxena, Kristina Lerman, and Aram Galstyan. 2022.
\newblock \href {https://arxiv.org/abs/1908.09635} {A survey on bias and fairness in machine learning}.
\newblock \emph{Preprint}, arXiv:1908.09635.

\bibitem[{{Mistral AI}(2024)}]{mistralai_mistral_large_2024}
{Mistral AI}. 2024.
\newblock \href {https://mistral.ai/news/mistral-large} {{Mistral Large}}.
\newblock \url{https://mistral.ai/news/mistral-large}.
\newblock Accessed: 2026-01-15.

\bibitem[{{OpenAI}(2024)}]{openai_gpt4_turbo_2024}
{OpenAI}. 2024.
\newblock \href {https://platform.openai.com/docs/models/gpt-4-turbo} {{GPT-4-Turbo}}.
\newblock \url{https://platform.openai.com/docs/models/gpt-4-turbo}.
\newblock Accessed: 2026-01-15.

\bibitem[{Papineni et~al.(2002)Papineni, Roukos, Ward, and Zhu}]{papineni2002bleu}
Kishore Papineni, Salim Roukos, Todd Ward, and Wei-Jing Zhu. 2002.
\newblock \href {https://doi.org/10.3115/1073083.1073135} {Bleu: a method for automatic evaluation of machine translation}.
\newblock In \emph{Proceedings of the 40th Annual Meeting on Association for Computational Linguistics}, ACL '02, page 311–318, USA. Association for Computational Linguistics.

\bibitem[{Rajpurkar et~al.(2016)Rajpurkar, Zhang, Lopyrev, and Liang}]{rajpurkar2016squad}
Pranav Rajpurkar, Jian Zhang, Konstantin Lopyrev, and Percy Liang. 2016.
\newblock \href {https://arxiv.org/abs/1606.05250} {Squad: 100,000+ questions for machine comprehension of text}.
\newblock \emph{Preprint}, arXiv:1606.05250.

\bibitem[{Wang et~al.(2023)Wang, Li, Chen, Cai, Zhu, Lin, Cao, Liu, Liu, and Sui}]{wang2023fair}
Peiyi Wang, Lei Li, Liang Chen, Zefan Cai, Dawei Zhu, Binghuai Lin, Yunbo Cao, Qi~Liu, Tianyu Liu, and Zhifang Sui. 2023.
\newblock \href {https://arxiv.org/abs/2305.17926} {Large language models are not fair evaluators}.
\newblock \emph{Preprint}, arXiv:2305.17926.

\bibitem[{Wang et~al.(2024)Wang, Yu, Zeng, Yang, Wang, Chen, Jiang, Xie, Wang, Xie, Ye, Zhang, and Zhang}]{wang2023pandalm}
Yidong Wang, Zhuohao Yu, Zhengran Zeng, Linyi Yang, Cunxiang Wang, Hao Chen, Chaoya Jiang, Rui Xie, Jindong Wang, Xing Xie, Wei Ye, Shikun Zhang, and Yue Zhang. 2024.
\newblock \href {https://arxiv.org/abs/2306.05087} {Pandalm: An automatic evaluation benchmark for llm instruction tuning optimization}.
\newblock \emph{Preprint}, arXiv:2306.05087.

\bibitem[{Yu et~al.(2024)Yu, Gao, Yao, Wang, Ye, Wang, Xie, Zhang, and Zhang}]{yu-etal-2024-kieval}
Zhuohao Yu, Chang Gao, Wenjin Yao, Yidong Wang, Wei Ye, Jindong Wang, Xing Xie, Yue Zhang, and Shikun Zhang. 2024.
\newblock \href {https://doi.org/10.18653/v1/2024.acl-long.325} {{KIE}val: A knowledge-grounded interactive evaluation framework for large language models}.
\newblock In \emph{Proceedings of the 62nd Annual Meeting of the Association for Computational Linguistics (Volume 1: Long Papers)}, pages 5967--5985, Bangkok, Thailand. Association for Computational Linguistics.

\bibitem[{Zhang et~al.(2025)Zhang, Cui, Chen, Wang, Zhang, Wang, Wu, and Hu}]{zhang2025stopovervaluingmultiagentdebate}
Hangfan Zhang, Zhiyao Cui, Jianhao Chen, Xinrun Wang, Qiaosheng Zhang, Zhen Wang, Dinghao Wu, and Shuyue Hu. 2025.
\newblock \href {https://arxiv.org/abs/2502.08788} {Stop overvaluing multi-agent debate -- we must rethink evaluation and embrace model heterogeneity}.
\newblock \emph{Preprint}, arXiv:2502.08788.

\bibitem[{Zheng et~al.(2023)Zheng, Chiang, Sheng, Zhuang, Wu, Zhuang, Lin, Li, Li, Xing, Zhang, Gonzalez, and Stoica}]{zheng2023judging}
Lianmin Zheng, Wei-Lin Chiang, Ying Sheng, Siyuan Zhuang, Zhanghao Wu, Yonghao Zhuang, Zi~Lin, Zhuohan Li, Dacheng Li, Eric~P. Xing, Hao Zhang, Joseph~E. Gonzalez, and Ion Stoica. 2023.
\newblock \href {https://arxiv.org/abs/2306.05685} {Judging llm-as-a-judge with mt-bench and chatbot arena}.
\newblock \emph{Preprint}, arXiv:2306.05685.

\bibitem[{Ziegler et~al.(2020)Ziegler, Stiennon, Wu, Brown, Radford, Amodei, Christiano, and Irving}]{ziegler2020finetuning}
Daniel~M. Ziegler, Nisan Stiennon, Jeffrey Wu, Tom~B. Brown, Alec Radford, Dario Amodei, Paul Christiano, and Geoffrey Irving. 2020.
\newblock \href {https://arxiv.org/abs/1909.08593} {Fine-tuning language models from human preferences}.
\newblock \emph{Preprint}, arXiv:1909.08593.

\end{thebibliography}

\appendix

\section{Note on AI assistance}
AI assistance was used for code development and improving the phrasing of the manuscript, while all analyses and conclusions were independently derived by the authors.

\section{Algorithms}
\begin{algorithm}
\caption{Multi-Advocate One-Round Evaluation (MORE)}
\begin{algorithmic}[1]
\State Initialize advocates $A_1 = \{A_{11},..., A_{1k}\}$ for Answer 1, $A_2 = \{A_{21},..., A_{2k}\}$ for Answer 2.
\State Initialize defenses $D_1 \leftarrow \emptyset$, $D_2 \leftarrow \emptyset$.
\For{$i=1$ to $k$} \Comment{Parallel argument generation}
    \State $d_{1i} \leftarrow \text{GenerateArgument}(A_{1i}, \text{Answer 1})$
    \State $D_1 \leftarrow D_1 \cup \{d_{1i}\}$
    \State $d_{2i} \leftarrow \text{GenerateArgument}(A_{2i}, \text{Answer 2})$
    \State $D_2 \leftarrow D_2 \cup \{d_{2i}\}$
\EndFor
\State $D_{1,agg} \leftarrow \text{AggregateDefenses}(D_1)$
\State $D_{2,agg} \leftarrow \text{AggregateDefenses}(D_2)$
\State Compile transcript $T$ with aggregated defenses.
\State $V \leftarrow \emptyset$ \Comment{Jury deliberation}
\For{each Juror $C_i$ in panel}
    \State $v_i \leftarrow \text{Vote}(C_i, T)$
    \State $V \leftarrow V \cup \{v_i\}$
\EndFor
\State winner $\leftarrow \text{AggregateVotes}(V)$ \Comment{Tie-break with Judge's score if needed}
\State \textbf{return} winner
\end{algorithmic}
\end{algorithm}

\begin{algorithm}
\caption{Single Advocate Multi-Round Evaluation (SAMRE) with Budgeted Stopping}
\begin{algorithmic}[1]
\State Initialize advocates $A_1, A_2$, Judge $J$, Jurors $\{C_1,..., C_m\}$, max rounds $R_{max}$, budget $B$.
\State Initialize transcript $T_0 \leftarrow \emptyset$, scores $S \leftarrow []$.
\For{$r=1$ to $R_{max}$}
    \State $d_1^r, d_2^r \leftarrow \text{GenerateArguments}(A_1, A_2, T_{r-1})$ \Comment{Advocates update arguments}
    \State $s_1^r, s_2^r, F^r \leftarrow \text{Evaluate}(J, d_1^r, d_2^r)$ \Comment{Judge scores and gives feedback}
    \State $S.\text{append}((s_1^r, s_2^r))$
    \State $T_r \leftarrow T_{r-1} \cup \{d_1^r, d_2^r, s_1^r, s_2^r, F^r\}$
    \If{$\text{CheckConvergence}(S, \epsilon)$ or $\text{TokenCost}(T_r) > B$}
        \State \textbf{break}
    \EndIf
\EndFor
\State $V \leftarrow \emptyset$ \Comment{Jury deliberation on final transcript}
\For{$i=1$ to $m$}
    \State $v_i \leftarrow \text{Vote}(C_i, T_r)$
    \State $V \leftarrow V \cup \{v_i\}$
\EndFor
\State winner $\leftarrow \text{AggregateVotes}(V)$ \Comment{Tie-break with Judge's final score}
\State \textbf{return} winner
\end{algorithmic}
\end{algorithm}

\section{Proofs}
\label{app:proofs}

\subsection{Proof of Theorem 1 (Probabilistic Convergence)}
\label{app:prob_convergence}
\begin{proof}
The theorem states that if $\lim_{r\to\infty} \mathbb{E}[\delta_r] = \Delta > 0$, then $\delta_r$ converges in probability to $\Delta$. We want to show $\lim_{r\to\infty} P(|\delta_r - \Delta| < \epsilon) = 1$ for any $\epsilon > 0$.

We use the triangle inequality: $|\delta_r - \Delta| \leq |\delta_r - \mathbb{E}[\delta_r]| + |\mathbb{E}[\delta_r] - \Delta|$.
For the event $\{|\delta_r - \Delta| \ge \epsilon\}$ to occur, it must be that either $\{|\delta_r - \mathbb{E}[\delta_r]| \ge \epsilon/2\}$ or $\{|\mathbb{E}[\delta_r] - \Delta| \ge \epsilon/2\}$.

By the assumption of convergence of the mean, for any $\epsilon > 0$, there exists an $N_1$ such that for all $r \ge N_1$, $|\mathbb{E}[\delta_r] - \Delta| < \epsilon/2$. So the second condition does not hold for large $r$.

Now consider the first condition. By Chebyshev's inequality:
\[
P(|\delta_r - \mathbb{E}[\delta_r]| \ge \epsilon/2) \le \frac{\text{Var}(\delta_r)}{(\epsilon/2)^2} = \frac{4\text{Var}(\delta_r)}{\epsilon^2}.
\]

The variance of the Beta posterior is $\text{Var}(\delta_r) = \frac{\alpha_r \beta_r}{(\alpha_r + \beta_r)^2 (\alpha_r + \beta_r + 1)}$. Since $\alpha_r + \beta_r = \alpha_0 + \beta_0 + r$, the denominator grows as $O(r^3)$, while the numerator $\alpha_r \beta_r$ grows at most as $O(r^2)$. Thus, $\text{Var}(\delta_r) = O(1/r)$, and $\lim_{r\to\infty} \text{Var}(\delta_r) = 0$.

Therefore, $\lim_{r\to\infty} P(|\delta_r - \mathbb{E}[\delta_r]| \ge \epsilon/2) = 0$.
Since both sources of deviation become arbitrarily small, $\lim_{r\to\infty} P(|\delta_r - \Delta| \ge \epsilon) = 0$, which completes the proof.
\end{proof}

\subsection{Proof of Theorem 2 (Score-Separation via Parallel Advocacy)}
\label{app:score_separation}
\begin{proof}
Let $g(f_{i,j})$ be the score of the $j$-th advocate for answer $a_i$. Let $G_i$ be the random variable representing the score of a single advocate for answer $a_i$. In the multi-advocate framework, the aggregated score is $M_i = \max(G_{i,1},..., G_{i,k})$.

We assume that answer $a_1$ is superior to $a_2$, formalized by stating that the cumulative distribution function (CDF) of $G_1$, denoted $F_1(x)$, first-order stochastically dominates (FOSD) the CDF of $G_2$, denoted $F_2(x)$. That is, $F_1(x) \le F_2(x)$ for all $x$, and the inequality is strict for some $x$. This implies $\mathbb{E}[G_1] > \mathbb{E}[G_2]$.

The CDF of the maximum of $k$ i.i.d. samples from $G_i$ is $F_{M_i}(x) = (F_i(x))^k$.
Since $F_1(x) \le F_2(x)$ for all $x$, it follows that $(F_1(x))^k \le (F_2(x))^k$. This means that $M_1$ also FOSD-dominates $M_2$, and thus $\mathbb{E}[M_1] > \mathbb{E}[M_2]$.

Furthermore, the operation of taking the maximum tends to stretch the upper tail of a distribution. The improvement from taking the maximum is expected to be greater for the stochastically larger distribution ($G_1$).
Formally, $\mathbb{E}[M_1] - \mathbb{E}[G_1] \ge \mathbb{E}[M_2] - \mathbb{E}[G_2]$.
This leads to a greater separation in expected scores:
\[
\mathbb{E}[M_1 - M_2] = \mathbb{E}[M_1] - \mathbb{E}[M_2] > \mathbb{E}[G_1] - \mathbb{E}[G_2].
\]
This completes the proof.
\end{proof}

\subsection{Proof of Theorem 3 (Score Differentiation)}
\label{app:score_differentiation}
\begin{proof}
Let $g(f_{i,j})$ be the score of the $j$-th advocate for answer $a_i$. Let $G_i$ be the random variable representing the score of a single advocate for answer $a_i$. In the multi-advocate framework, the aggregated score is $g(f_{i,agg}) = \max_{j} g(f_{i,j})$. Let $M_i = \max(G_{i,1},..., G_{i,k})$ be the random variable for the aggregated score.

We assume that answer $a_1$ is superior to $a_2$. This can be formalized by stating that the cumulative distribution function (CDF) of $G_1$, denoted $F_1(x)$, is stochastically smaller than the CDF of $G_2$, denoted $F_2(x)$. That is, $F_1(x) \le F_2(x)$ for all $x$, and there exists some $x$ for which the inequality is strict. This implies $\mathbb{E}[G_1] > \mathbb{E}[G_2]$.

The CDF of the maximum of $k$ i.i.d. samples from $G_i$ is $F_{M_i}(x) = (F_i(x))^k$.
Since $F_1(x) \le F_2(x)$, it follows that $(F_1(x))^k \le (F_2(x))^k$. This means that $M_1$ is also stochastically larger than $M_2$, and thus $\mathbb{E}[M_1] > \mathbb{E}[M_2]$.

Furthermore, the operation of taking the maximum tends to stretch the upper tail of a distribution. The difference between the expected value of the maximum of $k$ samples and the expected value of a single sample is larger for distributions with more mass in the upper tail. Because $G_1$ is stochastically larger than $G_2$, the improvement from taking the maximum is expected to be greater for $a_1$.
\[
\mathbb{E}[M_1] - \mathbb{E}[G_1] \ge \mathbb{E}[M_2] - \mathbb{E}[G_2].
\]
This leads to a greater separation in expected scores:
\[
\mathbb{E}[M_1 - M_2] = \mathbb{E}[M_1] - \mathbb{E}[M_2] > \mathbb{E}[G_1] - \mathbb{E}[G_2].
\]
This completes the proof.
\end{proof}

\section{Notation and Scoring Criteria}

\subsection{Notation}
\label{appendix:notation}
\begin{itemize}
    \item $A = \{A_1, A_2\}$: Set of advocates, where each advocate $A_i$ defends a specific answer.
    \item $J$: The judge who evaluates the arguments presented by the advocates.
    \item $C = \{C_1, C_2, C_3\}$: Set of jurors, where each juror $C_i$ casts a vote at the end of the evaluation process.
    \item $s_1^r$ and $s_2^r$: Scores given by the judge in the $r$-th round, corresponding to the evaluations of $A_1$ and $A_2$, respectively.
    \item $M_r$: The aggregated memory of all rounds up to the $r$-th round, which includes arguments, scores, and feedback.
    \item $f_A(A, M_{r-1})$: Function that generates the arguments $a_1^r$ and $a_2^r$ for the advocates based on the previous memory $M_{r-1}$.
    \item $f_J(J, a_1^r, a_2^r)$: Function that takes the judge and the arguments from the advocates, returning their scores $s_1^r$, $s_2^r$, and feedback $F^r$.
    \item $f_{C_i}(C_i, M_r)$: Function that represents the voting decision made by each juror $C_i$ based on the final memory $M_r$.
    \item $D_i$: The aggregated defense obtained by asking the LLM to consolidate the group's defenses into a single summary.
\end{itemize}

\subsection{Scoring Criteria}
\label{appendix:scoring_criteria}
The judge scores the advocates' arguments based on the following criteria, using a scale of 1-20:
\begin{itemize}
    \item Relevance to the question
    \item Accuracy of information and use of credible sources
    \item Depth of analysis and completeness of argument
    \item Clarity of expression and logical flow
    \item Strength of reasoning and factual support
    \item Effectiveness in addressing opponent's points
\end{itemize}

\subsection{Juror Backgrounds}
\label{appendix:juror}
In the SAMRE design, we selected jurors with varied professional backgrounds and perspectives:
\begin{itemize}
    \item A retired professor of ethics
    \item A young environmental activist
    \item A middle-aged business owner
    \item A social worker specializing in community development
    \item A technology entrepreneur with a background in AI
\end{itemize}

\section{Data Preprocessing and Evaluation}

\subsection{Artifact Licensing and Availability}
All benchmarks used in this study (MT-Bench, AlignBench, AUTO-J) are publicly available for research purposes under their respective licenses. Model APIs (GPT-4, Claude-3, Llama-3, Mistral) were accessed through their standard commercial or open-source terms of service. Baseline implementations follow the specifications in their original publications.

\subsection{Data Preprocessing}

To prepare the raw data for analysis, we implemented a script that processes the input data and generates an Excel file  structured with the following columns:

\begin{itemize}
    \item \textbf{Question}: This column contains the aggregated user questions used for evaluation.
    \item \textbf{Response\_A}: This column includes the responses generated by Model A for each corresponding question.
    \item \textbf{Response\_B}: This column presents the responses generated by Model B for the same set of questions.
    \item \textbf{Model\_A\_Score}: This binary score indicates the performance of Model A, where a score of 1 signifies a win and 0 signifies a loss in comparison to Model B.
    \item \textbf{Model\_B\_Score}: Similarly, this binary score reflects the performance of Model B, with a score of 1 representing a win and 0 representing a loss against Model A.
\end{itemize}

This structured format allows for straightforward analysis and comparison of the models' performances based on user questions and their respective responses.

\section{Agent Interaction Prompts}
\label{appendix:prompts}

We provide the detailed prompts used for the interactions between agents in our proposed architecture. The prompts are designed to guide the agents effectively throughout the evaluation process.

\subsection{Multi-Advocate One-Round Evaluation (MORE) Architecture Prompts}

\subsubsection{Judge Prompt}
\begin{tcolorbox}[width=\columnwidth, colback=gray!10, colframe=black, boxrule=0.5pt, left=2pt, right=2pt, top=2pt, bottom=2pt]
\small\ttfamily
You're a critical, impartial judge in a high-stakes debate on: "\{question\}". \\
Answer 1: "\{answer1\}". Answer 2: "\{answer2\}". \\
Your goal is to provide detailed, constructive feedback that will push advocates to significantly improve their arguments. \\
Current round: \{current\_round\} \\
Max rounds: \{max\_rounds\} \\
Previous scores: \{previous\_scores\} \\
\\
Defense for 1st answer: \{defense1\} \\
Defense for 2nd answer: \{defense2\} \\
\\
Analyze each argument meticulously. Be thorough and unbiased in your assessment of: \\
1. Relevance to the question \\
2. Accuracy of information and use of credible sources \\
3. Depth of analysis and completeness of argument \\
4. Clarity of expression and logical flow \\
5. Strength of reasoning and factual support \\
6. Effectiveness in addressing opponent's points \\
\\
For each criterion, provide a score on a scale of 1-20 and detailed justification. \\
Scores should be given as [Answer1\_score, Answer2\_score] for each criterion. \\
\\
Your comprehensive feedback for each advocate (50 words each): \\
Feedback for Advocate 1: \\
Feedback for Advocate 2: \\
\\
Sum up the scores and return the final score tuple (score1, score2). Example: (95, 87) \\
Your detailed scores and final tally:
\end{tcolorbox}

\subsubsection{Advocate Prompts}
\begin{tcolorbox}[width=\columnwidth, colback=gray!10, colframe=black, boxrule=0.5pt, left=2pt, right=2pt, top=2pt, bottom=2pt]
\small\ttfamily
You're a fierce advocate defending this answer: \{answer\} to the question: \{question\}. Your opponent defends: \{opponent\_answer\}. Convince the audience your argument is superior by addressing these criteria: \\
1. Relevance \\
2. Accuracy \\
3. Depth \\
4. Clarity \\
5. Logic and Factuality \\
\\
Be assertive and don't hold back! Use rhetoric and persuasion to win over the audience but be respectful. \\
Latest feedback: \{feedback\} \\
Opponent's last argument: \{opponent\_argument\} \\
\\
Your fierce defense (100 words max):
\end{tcolorbox}

\subsubsection{Summarizer Prompt}
\begin{tcolorbox}[width=\columnwidth, colback=gray!10, colframe=black, boxrule=0.5pt, left=2pt, right=2pt, top=2pt, bottom=2pt]
\small\ttfamily
Summarize the following content in 50 words or less, if there are any scores tuples, return them, it's important! Start summarization directly, no introductory sentences like here's your summary. In your summarization, only focus on the last scores, no partial ones. This is important: return the tuple of scores. These are the key points to summarize:\textbackslash n\textbackslash n\{content\}
\end{tcolorbox}

\subsection{Single Advocate Multi-Round Evaluation (SAMRE)}

\subsubsection{Defend Answer Prompt}
\begin{tcolorbox}[width=\columnwidth, colback=gray!10, colframe=black, boxrule=0.5pt, left=2pt, right=2pt, top=2pt, bottom=2pt]
\small\ttfamily
You're an eloquent advocate (Advocate \{advocate\_id\}) in a group defending this answer: \{answer\} to the question: \{question\}. Your opponent group defends: \{opponent\_answer\}. Collaborate with your fellow advocates to convince the audience your argument is better. Use the latest feedback, your opponent's last argument, and your team's previous arguments to improve your case. \\
Latest feedback: \{feedback\} \\
Opponent's last argument: \{opponent\_argument\} \\
Your team's previous arguments: \{team\_arguments\} \\
Respond in under 80 words. \\
Your defense:
\end{tcolorbox}

\subsubsection{Aggregate Defense Prompt}
\begin{tcolorbox}[width=\columnwidth, colback=gray!10, colframe=black, boxrule=0.5pt, left=2pt, right=2pt, top=2pt, bottom=2pt]
\small\ttfamily
You are an expert debate strategist. Your task is to aggregate and improve upon the following defenses for the answer: \{answer\} to the question: \{question\}. The opponent's answer is: \{opponent\_answer\}. \\
Individual defenses: \{defenses\} \\
Latest feedback from the judge: \{feedback\} \\
Analyze each defense critically. Identify the strongest points, address any weaknesses, and combine the best arguments into a cohesive, powerful defense. Aim to create a defense that is stronger and more comprehensive than any individual argument. \\
Provide your aggregated and improved defense in under 150 words:
\end{tcolorbox}

\subsubsection{Judge Answer Prompt}
\begin{tcolorbox}[width=\columnwidth, colback=gray!10, colframe=black, boxrule=0.5pt, left=2pt, right=2pt, top=2pt, bottom=2pt]
\small\ttfamily
You're a fair, impartial judge in a debate on: "\{question\}". Answer 1: "\{answer1\}". Answer 2: "\{answer2\}". Your goal is to provide feedback that will help advocate groups improve and differentiate their arguments more clearly. \\
Current round: \{current\_round\} \\
Total rounds: \{total\_rounds\} \\
Previous scores: \{previous\_scores\} \\
Defense for 1st answer: \{defense1\} \\
Defense for 2nd answer: \{defense2\} \\
Provide specific, constructive feedback to help each advocate group strengthen their unique position. Encourage them to address weaknesses and highlight distinctions. Aim for your feedback to lead to more divergent scores in future rounds. \\
Give your feedback in under 50 words:
\end{tcolorbox}

\subsubsection{Score Answer Prompt}

\begin{tcolorbox}[width=\columnwidth, colback=gray!10, colframe=black, boxrule=0.5pt, left=2pt, right=2pt, top=2pt, bottom=2pt]
\small\ttfamily
You're a critical, impartial judge in a high-stakes debate on: "\{question\}". Answer 1: "\{answer1\}". Answer 2: "\{answer2\}". Your goal is to provide detailed, constructive feedback that will push advocates to significantly improve their arguments. \\
Total rounds: \{total\_rounds\} \\
Previous scores: \{previous\_scores\} \\
Defense for 1st answer: \{defense1\} \\
Defense for 2nd answer: \{defense2\} \\
Analyze each argument meticulously. Be thorough and unbiased in your assessment of: \\
1. Relevance to the question \\
2. Accuracy of information and use of credible sources \\
3. Depth of analysis and completeness of argument \\
4. Clarity of expression and logical flow \\
5. Strength of reasoning and factual support \\
6. Effectiveness in addressing opponent's points \\
For each criterion, provide a score on a scale of 1-20 and detailed justification. Scores should be given as [Answer1\_score, Answer2\_score] for each criterion. \\
Your comprehensive feedback for each advocate (50 words each): \\
Feedback for Advocate 1: \\
Feedback for Advocate 2: \\
Sum up the scores and return the final score tuple (score1, score2). Example: (95, 87) \\
Your detailed scores and final tally:
\end{tcolorbox}

\subsection{Baseline Model Prompt}

\begin{tcolorbox}[width=\columnwidth, colback=gray!10, colframe=black, boxrule=0.5pt, left=2pt, right=2pt, top=2pt, bottom=2pt]
\small\ttfamily
You are a fair, impartial judge scoring a debate on the following question: \{question\}. \\
Answer 1: \{answer1\} \\
Answer 2: \{answer2\} \\
Score each answer on a scale of 1-20 for each of the following criteria: \\
  1. Relevance to the question \\
  2. Accuracy of information and use of credible sources \\
  3. Depth of analysis and completeness of argument \\
  4. Clarity of expression and logical flow \\
  5. Strength of reasoning and factual support \\
  6. Effectiveness in addressing opponent's points \\
Provide scores as [Answer1\_score, Answer2\_score] for each criterion in a list format, then sum for final scores. Please keep an eye on the slightest difference that should make a difference in the scoring. Don't overthink! \\
Relevance: \\
Accuracy: \\
Depth: \\
Clarity: \\
Logic and Factuality: \\
Addressing opponent's points: \\
Final Scores (sum of above) as a tuple (example: (18, 9)): \\
Explain your scoring, focusing on why one answer is better than the other based on the criteria above. Keep your explanation concise but informative. \\
Finally, return the final score tuple (score1, score2) as a tuple (in parentheses). Example: (18, 9) \\
Your scores and explanation:
\end{tcolorbox}

\section{Persona Sensitivity Analysis: Full Ablation}
\label{app:persona_ablation}

\subsection{Systematic Persona Pool and Robustness Analysis}
\label{subsec:persona_pool}

We constructed a diverse pool of 50 personas spanning law, medicine, education, technology, ethics, business, social work, risk analysis, and compliance. We then conducted 10 independent experiments on MT-Bench (120 randomly selected questions), each using a different random 5-persona subset. All personas were instantiated with GPT-4-Turbo as the backbone; personas influenced only instructional prompts, not underlying model capabilities. Table~\ref{tab:persona_robustness} summarizes results:

\begin{table*}[h!]
\centering
\caption{Persona Robustness Analysis: Ablation Across Random 5-Persona Subsets (MT-Bench, 120 questions)}
\label{tab:persona_robustness}
\begin{tabular}{lcc}
\toprule
\textbf{Configuration} & \textbf{Accuracy} (\%) \\
\midrule
Random 5-persona subsets (avg of 10) & $85.9 \pm 0.7$ \\
Curated 5-persona set & $86.1$ \\
Generic jurors (no personas) & $82.7 \pm 1.1$ \\
\bottomrule
\end{tabular}
\end{table*}

\paragraph{(1) Persona Conditioning Effect:} Persona-guided evaluation outperforms generic jurors by $+3.2\%$ (85.9\% vs.\ 82.7\%, $p<0.01$). The persona effect holds robustly across random subsets, indicating diversity, not specific persona identity, drives performance.

\paragraph{(2) Robustness Across Subsets:} Tight variance ($0.7\%$ std.\ dev.) indicates D3's performance is not brittle to persona composition; the framework generalizes well to unseen persona combinations.

\paragraph{(3) Curated vs.~Random:} Curated personas ($86.1\%$) achieve marginally higher accuracy than the random average ($85.9\%$, $+0.2\%$). This validates that deliberate design adds value without harming generalization.

\subsection{Design Rationale: Persona Selection and Value Lenses}
\label{subsec:persona_rationale}

Our five curated personas were selected to provide complementary professional perspectives and value lenses. Each persona is anchored in domain expertise rather than demographic attributes, reducing stereotype risks:

\begin{table*}[h]
\centering
\begin{tabular}{l l c c}
\toprule
\textbf{Framework} & \textbf{Evaluator} & \textbf{Accuracy (\%)} & \textbf{Relative Gain} \\
\midrule
Single Judge & Llama-3-70B & 68.3 & baseline \\
ChatEval & Llama-3-70B & 73.1 & +4.8\% \\
PRD & Llama-3-70B & 71.5 & +3.2\% \\
D3-MORE & Llama-3-70B & 73.9 & +5.6\% \\
\midrule
D3-MORE & GPT-4-Turbo & 85.1 & +12.6\% \\
\bottomrule
\end{tabular}
\caption{MT-Bench accuracy using open-source and proprietary evaluators.}
\label{tab:opensource_mtbench}
\end{table*}

\begin{itemize}
  \item \textbf{Ethics Professor:} Ethical principles, long-term societal impact
  \item \textbf{Environmental Activist:} Collective welfare, ecological responsibility
  \item \textbf{Business Owner:} Practical feasibility, \textsc{roi}, trade-offs
  \item \textbf{Social Worker:} Human-centered perspective, equity, vulnerable populations
  \item \textbf{Tech Entrepreneur:} Innovation, scalability, technological progress
\end{itemize}

This composition ensures verdicts incorporate ethical grounding, distributional impact, economic viability, and technological feasibility. These roles are intentionally role-based rather than demographic; all jurors execute the same backbone model with identical base capabilities.

\subsection{Cross-Persona Agreement and Complementarity}
\label{subsec:persona_agreement}

We computed pairwise vote agreement across the five curated personas. Average agreement is $89.5\% \pm 1.5\%$, with pairwise ranges $87{-}92\%$. High cross-persona agreement confirms personas are complementary rather than contradictory, reducing the risk of arbitrarily conflicting judgments while maintaining diverse perspectives.

\subsection{SAMRE Debate Progression Across Question Types}
\label{app:samre-progression}

The convergence behavior of iterative debate varies systematically across task complexity and reasoning demands. We analyze score gap trajectories across MT-Bench question categories to validate that the budgeted stopping rule terminates debates at their point of maximal signal without sacrificing verdict reliability.

Coding tasks exhibit the largest discriminative gaps, with peaks reaching 20 points by round 2 and maintaining plateau stability through round 5. This reflects clear correctness boundaries in code evaluation. Reasoning and ethics questions show moderate, steady gaps (8--11 points) with gentle convergence, indicating these subjective domains benefit from iterative refinement but stabilize predictably. Writing, roleplay, and math tasks display tighter gaps (0--6 points) and earlier plateau behavior, suggesting these domains reach verdict saturation with fewer debate rounds.

Across all categories, gap stabilization occurs by rounds 4--5, with the majority of verdicts determined by round 2. The final round markers (diamonds) per question type show that D3-SAMRE can terminate heterogeneously: early for low-variance tasks like math and writing, while extending slightly longer for nuanced reasoning and ethics evaluations. This task-sensitive convergence supports the empirical guidance in Section~\ref{sec:protocol_selection}: use SAMRE for high-stakes and ethically complex evaluations, while MORE suffices for objective or well-separated task types.

\begin{figure}[h]
	\centering
	\includegraphics[width=\columnwidth]{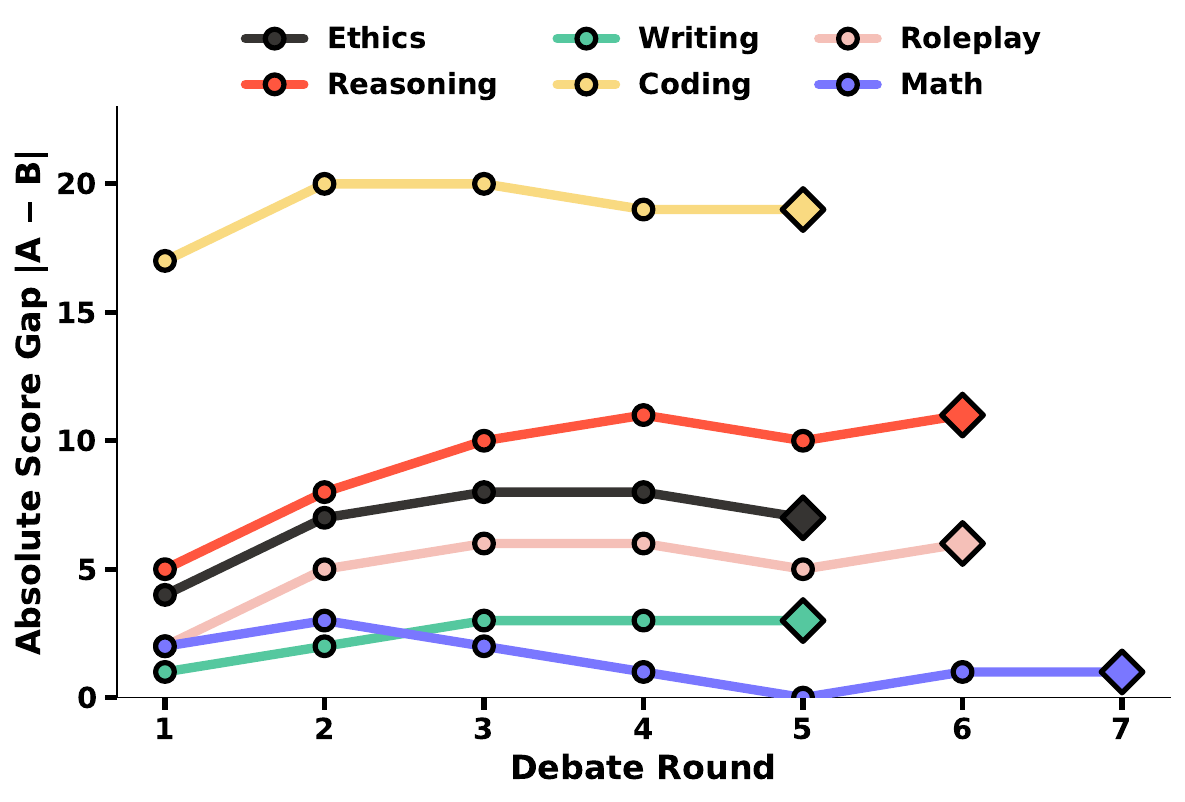}
	\caption{Score gap trajectories across six MT-Bench question categories reveal differential convergence patterns in iterative adversarial debate. Coding tasks exhibit the largest gaps (peak 20 points) with stable plateaus, indicating clear discriminability. Reasoning and ethics tasks show moderate, steady gaps (8--11 points). Writing, roleplay, and math questions display tighter gaps (0--6 points) with earlier plateau behavior. Diamond markers indicate final verdict round per category.}
	\label{fig:samre-progression}
\end{figure}

\section{Open-Source Evaluator Results}
\label{app:opensource}

To test whether D3’s performance gains depend on evaluator model quality, we evaluate all frameworks using \textbf{Llama-3-70B} \cite{grattafiori2024llama3herdmodels} as the evaluator on 100 MT-Bench \cite{zheng2023judging} questions. Table~\ref{tab:opensource_mtbench} reports accuracy and relative gains over a single-judge baseline.

These results show that D3’s improvements persist when using a fully open-source evaluator, indicating that the gains stem from the evaluation architecture rather than reliance on proprietary judge models.

\end{document}